\documentclass[10pt]{iopart}

\usepackage[utf8]{inputenc}
\expandafter\let\csname equation*\endcsname\relax
\expandafter\let\csname endequation*\endcsname\relax
\usepackage{amsmath}
\usepackage{graphicx}
\usepackage{amssymb}
\usepackage{amsfonts}
\usepackage{textcomp}
\usepackage{stfloats}
\usepackage{url}
\usepackage{verbatim}
\usepackage[ruled,vlined,linesnumbered]{algorithm2e}
\SetKwInput{KwInput}{Input}
\SetKwInput{KwOutput}{Output}

\usepackage{tikz}
\tikzstyle{arrow}=[->] %
\usetikzlibrary{calc}
\usetikzlibrary{bayesnet}
\usetikzlibrary{shapes.geometric}
\usetikzlibrary{shadows} 
\usetikzlibrary{backgrounds}
\usepackage{mathtools}
\usepackage{multirow}
\usepackage{hyperref}
\hypersetup{%
  colorlinks=false,%
  linkbordercolor=red,%
  pdfborderstyle={/S/U/W 1}%
}
\usepackage{caption}
\usepackage{subcaption}
\usepackage[makeroom]{cancel}
\usepackage{multirow}
\usepackage[sorting=none]{biblatex}
\DeclareMathOperator*{\argmin}{arg\,min}

\DeclareMathOperator*{\E}{\mathbb{E}}

\newcommand*{\lossCE}{\ensuremath{\mathcal{L}_{\textsc{CE}}}}
\newcommand*{\lossCensor}{\ensuremath{\mathcal{L}_{\textsc{censor}}}}

\DeclarePairedDelimiterX{\infdivx}[2]{\big(}{\big)}{%
  #1\;\delimsize\|\;#2%
}

\newcommand{\kldiv}{\ensuremath{\textrm{KL}\infdivx}}

\newcommand*\wc{{\mkern 2mu\cdot\mkern 2mu}}

\newcommand{\shortminus}{%
    \scalebox{0.75}[1.0]{\ensuremath{-}}
}

\begin{document}
\title{Stabilizing Subject Transfer in EEG Classification with Divergence Estimation}

\author{
 Niklas Smedemark-Margulies$^{1}$\footnote{Work done while NSM and YB were interns at MERL.},
 Ye Wang$^{2}$,
 Toshiaki Koike-Akino$^{2}$,
 Jing Liu$^{2}$,
 Kieran Parsons$^{2}$,
 Yunus Bicer$^{3}$,
 Deniz Erdo{\u{g}}mu{\c{s}}$^{3}$
 }
 \address{$^1$Khoury College of Computer Sciences, Northeastern University, Boston, MA, USA}
 \address{$^2$Mitsubishi Electric Research Labs. (MERL), Cambridge, MA, USA}
 \address{$^3$Department of Electrical and Computer Engineering, Northeastern University, Boston, MA, USA}
 
\begin{abstract}
\textit{Objective}. Classification models for electroencephalogram (EEG) data show a large decrease in performance when evaluated on unseen test subjects. We reduce this performance decrease using new regularization techniques during model training.
\textit{Approach}. 
We propose several graphical models to describe an EEG classification task. 
From each model, we identify statistical relationships that should hold true in an idealized training scenario (with infinite data and a globally-optimal model) but that may not hold in practice.
We design regularization penalties to enforce these relationships in two stages.
First, we identify suitable proxy quantities (divergences such as Mutual Information and Wasserstein-1) that can be used to measure statistical independence and dependence relationships. 
Second, we provide algorithms to efficiently estimate these quantities during training using secondary neural network models.
\textit{Main Results}. 
We conduct extensive computational experiments using a large benchmark EEG dataset, comparing our proposed techniques with a baseline method that uses an adversarial classifier.
We find our proposed methods significantly increase balanced accuracy on test subjects and decrease overfitting. 
The proposed methods exhibit a larger benefit over a greater range of hyperparameters than the baseline method, with only a small computational cost at training time.
These benefits are largest when used for a fixed training period, though there is still a significant benefit for a subset of hyperparameters when our techniques are used in conjunction with early stopping regularization. 
\textit{Significance}.
The high variability in signal structure between subjects means that typical approaches to EEG signal modeling often require time-intensive calibration for each user, and even re-calibration before every use. 
By improving the performance of population models in the most stringent case of zero-shot subject transfer, we may help reduce or eliminate the need for model calibration. Our results may also provide a beneficial starting point when used in combination with fine-tuning techniques.
\end{abstract}


\submitto{\JNE}
\maketitle
\ioptwocol

\section{Introduction}
In the field of signal modeling for electroencephalogram (EEG) and related biosignals, a key challenge is to train models that can extrapolate to unseen test subjects. 
It has been repeatedly observed in the literature~\cite{wu2020transfer} that signal models do not readily transfer to new subjects.
Multiple factors contribute to this performance gap, including data noise (due to the limits of sensor technology and signal attenuation between the brain surface and the scalp), 
label noise (due to the challenge of precisely adhering to cues in brain-computer interface experiments), and
intrinsic differences in signal structure across subjects.

We introduce two new regularization methods to reduce this performance gap during subject transfer. 
Our methods are based on a pre-existing framework for subject transfer learning known as ``censoring''~\cite{wang2018invariant}.
While the benefits of the censoring framework have been demonstrated empirically in previous research, we provide new theoretical motivation, as well as new implementations that are simple and effective across a wide range of hyperparameters. 
To derive a particular regularization penalty, we first select a generative model for the task and examine its conditional independence structure. We choose a statistical relationship that should hold true in an idealized classifier trained using data from this generative model, but which may not hold true in practice. 
We then convert this relationship to a regularization term by identifying a suitable quantity (a divergence such as mutual information or Wasserstein distance) to measure the relationship, and defining a simple algorithm for estimating this quantity during classifier training.
By enforcing these relationships, censoring helps classifiers converge with less overfitting, despite being trained on a finite, noisy sample of data.

\paragraph{High-level Approach.}
We first describe several possible generative models for an EEG classification task. 
For each generative model, we identify crucial statistical relationships that should hold in the limit of infinite data, but which may be violated in a finite training data sample.
We then identify surrogate quantities that can measure these statistical relationships, and provide estimation algorithms for these quantities that can be used during training.
The estimation algorithms we provide require limited additional resources during training.

\paragraph{Experiments.} We conduct extensive cross-validation experiments on a large benchmark EEG dataset to evaluate the effect of the proposed regularization methods. 
This benchmark dataset consists of binary EEG responses collected during a rapid serial visual presentation (RSVP) paradigm~\cite{zhang2020benchmark}.
In each experiment, we train an EEG classifier model on a subset of subjects, with or without regularization, and measure the model's balanced accuracy on a set of unseen test subjects.
To make a thorough statistical evaluation of our proposed methods, we perform over $60$K such experiments, varying hyperparameters such as the regularization penalty, model structure, as well the set of training, validation, and test subjects, and the random initialization of the model.

Our primary focus is increasing the model's test performance at the end of a fixed number of epochs, since this gives a direct comparison between a regularized and unregularized model.
We also include experiments measuring test performance at the epoch of best validation accuracy; these secondary experiments evaluate how our techniques work in combination with early stopping. 
Note that early stopping based on validation performance requires sacrificing a portion of training data, and may not be applicable in some settings.

\paragraph{Results.} We find that our method significantly improves balanced accuracy on the unseen test subjects, and also significantly reduces model overfitting. 
These benefits are most pronounced when measured after training for a fixed number of epochs, but still significant even when training is stopped early using validation metrics, indicating that our method provides regularization that is partially separate from the effect of early stopping.

\paragraph{Contributions.} The overall contributions of this work are as follows.
\begin{itemize}
    \item We provide a novel theoretical motivation for a range of censoring regularization penalties.
    \item We derive two simple and efficient new estimation techniques for enforcing these regularization penalties, based on density ratio estimation and Wasserstein distances.
    \item Using extensive computational experiments, we find that our proposed techniques significantly increase test performance and reduce overfitting. These benefits are larger and occur for a wider range of hyperparameters than a widely-studied baseline method.
\end{itemize}

\subsection{Related Work}
Brain-computer interface research often focuses on restoring communication in individuals with severe speech and physical impairment (SSPI).
Non-invasive electroencephalography (EEG) is a well-established modality for this purpose, with a wide variety of established experimental paradigms.

In query-and-response paradigms, a subject is queried with a stimulus (such as images on a screen) and their EEG response is measured. 
In particular, we focus on a paradigm called rapid serial visual presentation (RSVP)~\cite{lees2018review}. 
Briefly, a subject first imagines a target item from a pre-defined set, such as one letter of the alphabet.
The subject is queried with a sequence of multiple images in quick succession; each image in the sequence constitutes a binary trial, and contains one possible item from the pre-defined set.
The subject's EEG response to each trial provides evidence about which symbol is desired. 
A symbol may be selected from one trial or query sequence, or the evidence from multiple sequences can be accumulated to perform recursive Bayesian inference~\cite{smedemark2023recursive}.

EEG is used for numerous other communication paradigms, including other query-and-response methods such as steady-state visually-evoked potentials (SSVEP)~\cite{norcia2015steady}, and paradigms without a stimulus prompt such as motor imagery (MI)~\cite{wierzgala2018most} or classification of emotional affect~\cite{torres2020eeg}.
Subject transfer learning is a common challenge across these communication paradigms and for the modeling of related biosignals data types such as electromyography (EMG) and electrocorticography (ECoG)~\cite{jayaram2016transfer}.

Some work on subject transfer learning has applied domain adaptation methods, with the goal of harmonizing datasets from different subjects, measurement devices, or experimental paradigms. The goal in these approaches is to be able to train a single model on these collected datasets~\cite{congedo2017riemannian,liu2021align,zheng2016personalizing}. 

Other work has investigated the use of censoring penalties in training variational autoencoders~\cite{ozdenizci2019transfer} and learning disentangled representations\cite{han2020disentangled}. Other work has applied censoring penalties to enforce different notions of conditional independence, using estimation techniques such as kernel density estimation and neural critic functions~\cite{smedemark2022autotransfer}. Our work extends these approaches by providing a stronger theoretical motivation for each censoring penalty and providing two new methods for estimating censoring penalties that are highly effective and simple to implement.

The estimation techniques we develop here rely on several techniques from the generative modeling literature. One technique uses density ratio estimation~\cite{sugiyama2010density} to approximately compute a Mutual Information (MI) term; a similar technique has been previously demonstrated for other applications~\cite{suzuki2008approximating}. More recent work has explored other approaches to estimating MI~\cite{poole2019variational}. Our other technique replaces the use of Kullback-Leibler (KL) divergence with Wasserstein-1 distance in order to estimate dependence between variables. This technique has been previously described as a Wasserstein dependency measure~\cite{ozair2019wasserstein}; our approach to computing an estimate of the Wasserstein-1 distance is based on previous research on sampling realistic images~\cite{wasserstein-gan}.

\section{Methods}
\paragraph{Overview.} Here, we define the unseen subject classification task and motivate our approach. 
We provide three generative models describing this task. 
From these generative models, we select one or more statistical relationships at a time to enforce during model training for regularization; we refer to each choice of one relationship as a ``censoring mode.''
We formally define the components of our model architecture.
Next, we introduce several estimation techniques for measuring the statistical relationships that we hope to enforce, and show how to train our model with the desired regularization.
Finally, we describe the computational experiments that we perform to evaluate our proposed methods.

\subsection{Problem Statement and Motivation}
\label{sec:motivation}
Consider a dataset of tuples $\{(x, y, s)\}$, with data $x \in \mathbb{R}^D$, discrete task labels $y \in \{1, \ldots, C\}$, and discrete nuisance labels $s \in \{1, \ldots, S\}$. 
The nuisance labels represent the combination of subject identifier and session identifier. 
These tuples will be sampled from an empirical data distribution $(x, y, s) \sim p(X, Y, S)$, whose generative model is described below.
We seek to train a classifier on a subset of subjects, and regularize the model's training to achieve high accuracy on unseen test subjects.  
At test time, we will receive only a set of data $X$ from the test subject, and must infer the corresponding set of task labels $Y$.

\paragraph{Idealized and Real-world Settings.}
In order to train a classifier to infer $p(Y|X)$, we can first choose a generative model describing how we believe the dataset was produced.
If this generative model matches the true generating process for the dataset, and if training results in a classifier that is well-fit to the dataset, then we would expect to find that the trained classifier exhibits the same statistical relationships that exist in the generative model. 
For example, we would expect that variables which are independent in the generative model are also independent in the distribution learned by the classifier.
In an idealized setting where we have infinite, unbiased training data and a global optimization algorithm, we may expect this favorable outcome (where the learned model matches the generative model) with no additional effort.

In practice, however, we typically encounter several key limitations. 
Tasks involving biosignals such as EEG often have very limited training data that is both noisy and may come from a non-representative sample of subjects.
Furthermore, typical classifiers are trained using local optimization strategies, such as using stochastic gradient descent on a non-convex loss function. 
Thus, we do not expect models trained using only a classification objective to necessarily obey the correct dependence structure. 
In particular, note that models for biosignals classification tasks may incorrectly learn a distribution of features that correlates strongly with the subject identifier~\cite{ozdenizci2019adversarial}; essentially a form of overfitting to the training set. 
This may explain the common experimental observation of a ``subject transfer gap'' - a large decrease in model performance when tested on unseen subjects~\cite{wan2021review}. 

We reduce this subject transfer gap using regularization penalties.
By specifying a certain generative model, we have also implicitly defined a set of statistical relationships such as conditional independences. 
We can easily enumerate these relationships, e.g. using the ``Bayes Ball'' algorithm~\cite{shachter2013bayes}. 
For a pair of variables $A$ and $B$, conditioned on a set of zero or more additional observed variables $C$, we may identify that our model implies relationships such as a marginal independence $A \perp B$, a conditional independence $A \perp B | C$, or a conditional dependence $A \not\perp B | C$. 
Note that the set of all such statements is combinatorially large in the number of individual variables of the generative model; thus it is not feasible to enforce them all.
We select just one or two of these statistical relationships at a time, and enforce them as a regularization objective.
This approach helps the model converge to a better optimum that will generalize to unseen subjects with less overfitting.
We refer to these regularization objectives as ``censoring'' objectives, because we deliberately choose relationships involving the model's latent features and the nuisance labels.

\subsection{Graphical Models and Censoring modes.}
\label{sec:graphical-models-censor-modes}
\begin{figure}[tb]
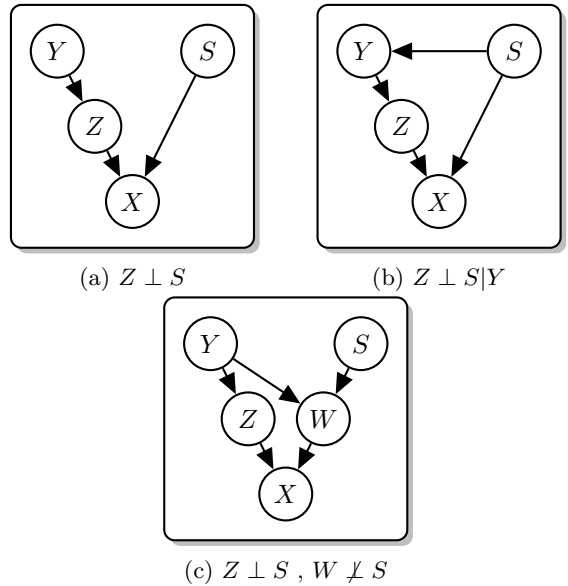

    \centering
    \begin{subfigure}[b]{0.49\linewidth}
    \centering
    \tikz{
    \node[latent, thick] (Y) at (0, 0) {$Y$};
    \node[latent, thick] (S) at (2, 0) {$S$};
    \node[latent, thick] (Z) at (0.5, -1) {$Z$};
    \node[latent, thick] (X) at (1, -2) {$X$};
    \begin{scope}[on background layer]
        \node[fill=white, draw, rounded corners, inner sep=0.25cm, thick, drop shadow, 
        fit=(Y) (S) (Z) (X)
    ] {};
    \end{scope}
    \draw[style=arrow, thick] (S) to (X);
    \draw[style=arrow, thick] (Y) to (Z);
    \draw[style=arrow, thick] (Z) to (X);
    }
    \caption{$Z\perp S$}
    \label{fig:graphical-model-1}
    \end{subfigure}%
    \begin{subfigure}[b]{0.49\linewidth}
    \centering
    \tikz{
    \node[latent, thick] (Y) at (0, 0) {$Y$};
    \node[latent, thick] (S) at (2, 0) {$S$};
    \node[latent, thick] (Z) at (0.5, -1) {$Z$};
    \node[latent, thick] (X) at (1, -2) {$X$};
    \begin{scope}[on background layer]
        \node[fill=white, draw, rounded corners, inner sep=0.25cm, thick, drop shadow, 
        fit=(Y) (S) (Z) (X)
    ] {};
    \end{scope}
    \draw[style=arrow, thick] (S) to (X);
    \draw[style=arrow, thick] (S) to (Y);
    \draw[style=arrow, thick] (Y) to (Z);
    \draw[style=arrow, thick] (Z) to (X);
    }
    \caption{$Z\perp S | Y$}
    \label{fig:graphical-model-2}
    \end{subfigure}
    \begin{subfigure}[b]{0.55\linewidth}
    \centering
    \tikz{
    \node[latent, thick] (Y) at (0, 0) {$Y$};
    \node[latent, thick] (S) at (2, 0) {$S$};
    \node[latent, thick] (Z) at (0.5, -1) {$Z$};
    \node[latent, thick] (W) at (1.5, -1) {$W$};
    \node[latent, thick] (X) at (1, -2) {$X$};
    \draw [style=arrow, thick] (Y) to (Z);
    \draw [style=arrow, thick] (Y) to (W);
    \draw [style=arrow, thick] (Z) to (X);
    \draw [style=arrow, thick] (S) to (W);
    \draw [style=arrow, thick] (W) to (X);
    \begin{scope}[on background layer]
        \node[fill=white, draw, rounded corners, inner sep=0.25cm, thick, drop shadow, 
        fit=(Y) (Z) (S) (W) (X)
    ] {};
    \end{scope}
    }
    \caption{$Z\perp S$ , $W \not\perp S$}
    \label{fig:graphical-model-3}
    \end{subfigure}
    \caption{Graphical models for EEG classification that motivate different regularization approaches.
    \subref{fig:graphical-model-1}: the distribution of actions does not differ across subjects $p(Y|S) = p(Y)$; introducing a latent variable $Z$ facilitates regularization by enforcing \textit{marginal} independence $Z \perp S$.
    \subref{fig:graphical-model-2}: actions may vary across subjects $p(Y|S) \neq p(Y)$; this correlation suggests enforcing \textit{conditional} independence $Z \perp S | Y$.
    \subref{fig:graphical-model-3}: a second latent variable is introduced to capture nuisance-related information for use inferring task labels; \textit{complementary} regularization is performed with a pair of penalties to enforce independence $Z \perp S$ and dependence $W \not\perp S$.}
    \label{fig:graphical-models}
\end{figure}

Figure~\ref{fig:graphical-models} shows three possible graphical models for an EEG classification task, each of which motivates a different regularization strategy. 

In Figure~\ref{fig:graphical-model-1}, we consider the case of a single latent variable $Z$ and define the generative process as 
\begin{align}
p(X, Y, Z) = p(S) p(Y) p(Z|Y) p(X|S,Z).
\end{align}
Here, the latent variable should be marginally independent of the nuisance labels $Z \perp S$, giving the first censoring mode which we refer to as \textbf{marginal censoring}. 
This model makes the simplifying assumption that the distribution of task labels does not differ across different subjects or sessions, so that there is no direct link between $S$ and $Y$ (i.e. $p(Y|S) = p(Y)$).

Figure~\ref{fig:graphical-model-2} relaxes this assumption and adds a connection from $S$ to $Y$; the resulting generative process is defined as
\begin{align}
p(X, Y, Z) = p(S) p(Y|S) p(Z|Y) p(X|S,Z).
\end{align}
This dependence could arise in an EEG typing task where a subject tends to use their preferred letters or words with higher frequency.
The connection between $S$ and $Y$ means that the latent variable is no longer marginally independent of the nuisance variable; we instead enforce conditional independence $Z \perp S | Y$, giving our second censoring mode called \textbf{conditional censoring}. 
Intuitively, this allows the latent features to have some information about the nuisance variable, but no more than the amount already implied by $Y$.

In Figure~\ref{fig:graphical-model-3}, to address the possibility that the nuisance variable may be informative when predicting the task label at test time, we include a second latent variable $W$ that captures nuisance-related information.
The generative process becomes
\begin{align}
p(X, Y, Z) = p(S) p(Y) p(Z|Y) p(W|S) p(X|Z,W).
\end{align}
Recall that for the unseen test subjects, the value of $S$ will not be available.
Furthermore, its value would not be directly useful to the model, since it comes from a region of the domain of $S$ that was never observed during training.
Instead, we hope to infer the second latent variable $W$, including some nuisance-related information, from the data $X$; this may help the classifier model to better predict $Y$.
In this model, one latent variable is marginally independent of the nuisance variable $Z \perp S$, while the other is strongly determined by the nuisance variable, which we merely describe as $W \not \perp S$ (note that we try to maximize this dependence in our penalties, even though this notation requires only a minimal correlation).
This censoring mode is called \textbf{complementary censoring}. 

\subsection{Model Architecture}
\definecolor{viridis_blue}{rgb}{0.22, 0.34, 0.55}
\definecolor{viridis_green}{rgb}{0.45, 0.82, 0.33}
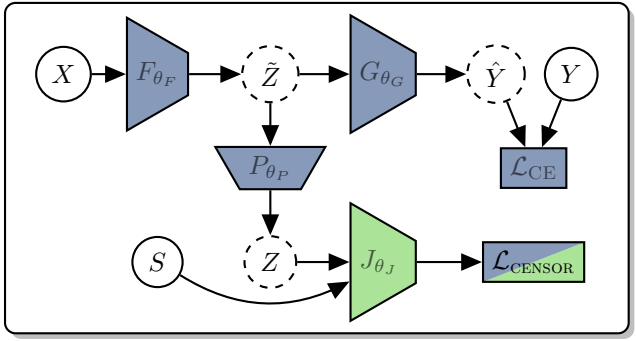
\begin{figure}[tb]
    \centering
    \tikzset{main_model/.style={trapezium,draw=black,text=black, thick, fill=viridis_blue, fill opacity=0.6, shape border rotate=270}}
    \tikzset{censor_model/.style={trapezium,draw=black,text=black, thick, fill=viridis_green, fill opacity=0.6, shape border rotate=270}}
    \tikzset{input var/.style={circle,draw=black,text=black, thick}}
    \tikzset{model var/.style={circle,draw=black,text=black, thick, dashed}}
    \tikzset{loss onecolor/.style={rectangle,draw=black,text=black, thick, fill=viridis_blue, fill opacity=0.6}}
    \tikzset{
        loss twocolor/.style={
            rectangle,draw=black,text=black, thick,
            path picture={
                \fill[viridis_green, fill opacity=0.6] (path picture bounding box.south west) -- (path picture bounding box.north east) |-cycle;
                \fill[viridis_blue, fill opacity=0.6] (path picture bounding box.north east) -- (path picture bounding box.south west) |-cycle;
            }
        }
    }
    \begin{tikzpicture}
        \node[input var] (x) at (0, 0) {$X$};
        \node[input var] (s) at (1.25, -2.5) {$S$};
        \node[main_model] (enc) at (1.25, 0) {$F_{\theta_F}$};
        \node[model var] (z_hidden) at (2.75, 0) {$\tilde{Z}$};
        \node[main_model] (clf) at (4.25, 0) {$G_{\theta_G}$};
        \node[model var] (y_hat) at (5.75, 0) {$\hat{Y}$};
        \node[input var] (y_true) at (6.75, 0) {$Y$};
        \node[main_model, shape border rotate=180] (proj) at (2.75, -1.25) {$P_{\theta_P}$};
        \node[model var] (z_obs) at (2.75, -2.5) {$Z$};
        \node[censor_model] (censor) at (4.25, -2.5) {$J_{\theta_J}\ $};
        \node[loss onecolor] (cross_ent) at (6.25, -1.25) {$\lossCE$};
        \node[loss twocolor] (censor_penalty) at (6.25, -2.5) {$\lossCensor$};
        \draw [style=arrow,thick] (x) to (enc);
        \draw [style=arrow,thick, bend right=30] (s) to (censor);
        \draw [style=arrow,thick] (enc) to (z_hidden);
        \draw [style=arrow,thick] (z_hidden) to (clf);
        \draw [style=arrow,thick] (clf) to (y_hat);
        \draw [style=arrow,thick] (y_hat) to (cross_ent);
        \draw [style=arrow,thick] (y_true) to (cross_ent);
        \draw [style=arrow,thick] (z_hidden) to (proj);
        \draw [style=arrow,thick] (proj) to (z_obs);
        \draw [style=arrow,thick] (z_obs) to (censor);
        \draw [style=arrow,thick] (censor) to (censor_penalty);
        \begin{scope}[on background layer]
            \node[fill=white, draw, rounded corners, inner sep=0.4cm, thick, drop shadow, 
            fit=
            (x) (enc) (z_hidden) (clf) (y_hat) (y_true) (proj) (z_obs) (censor) (censor_penalty) (cross_ent),
        ] {};
        \end{scope}
    \end{tikzpicture}
    \caption{Model Architecture. 
    Trapezoids are trainable models: encoder $F_{\theta_F}$, classifier $G_{\theta_G}$, projection $P_{\theta_P}$, and censoring model $J_{\theta_J}$. 
    Solid circles are input variables: data $X$, true task labels $Y$, and nuisance labels $S$.
    Dashed circles are intermediate variables: hidden features $\tilde{Z}$, observed features $Z$, and predicted task labels $\hat{Y}$. 
    Rectangles are loss terms: cross-entropy loss $\lossCE$ and regularization penalty $\lossCensor$.
    Training alternates between updating blue and green model components; both $\lossCE$ and $\lossCensor$ are used to update the main model, while the censoring model is only trained using $\lossCensor$ (with appropriate changes such as inverted sign; see below).
    $J_{\theta_J}$ receives additional inputs in some settings.
    }
    \label{fig:model-schematic}
\end{figure}
To approach the unseen subject classification task, we construct a task model to classify data and a censoring model to regularize the task model, as shown in Figure~\ref{fig:model-schematic}. 

For convenience, let $p(Z,Y,S)$ denote the distribution obtained by sampling from the empirical data distribution $(x, y, s) \sim p(X,Y,S)$ and then applying the encoder and projector $z = P_{\theta_P}(F_{\theta_F}(x))$.
Define $p(Z,S)$ as the same pushforward distribution after marginalizing over $Y$ (easily achieved by dropping $y$ after sampling); likewise define $p(Z)$ by marginalizing over both $Y$ and $S$.

The task model consists of an encoder $F_{\theta_F}(\wc): \mathbb{R}^D \to \mathbb{R}^K$ that produces $K$-dimensional \textit{hidden} features $\tilde{z} \in \mathbb{R}^K$, and a classifier $G_{\theta_G}(\wc): \mathbb{R}^K \to \Delta(C)$ that maps feature vectors to a vector in the $C$-dimensional probability simplex. 
The task model is trained so that predicted label distribution $\hat{y} = G_{\theta_G}(F_{\theta_F}(x)), x \sim p(X)$ approximates the true posterior distribution over labels $y \sim p(Y|X)$.
The projection $P_{\theta_P}(\wc): \mathbb{R}^K \to \mathbb{R}^K$ maps \textit{hidden} feature vectors to \textit{observed} feature vectors $\tilde{z} \mapsto z$;
this is included based on empirical benefits observed in the contrastive learning literature~\cite{gupta2022understanding}. In some experiments, this projection is the identity mapping with zero parameters, which we refer to as ``trivial'' or ``direct features''; otherwise, the projection is ``non-trivial'' and gives ``projected features.'' During training, the projection is updated along with the task model.

The censoring model $J_{\theta_J}(\wc): \mathbb{R}^T \to \mathbb{R}^L$ regularizes the task model. The censoring model's input and output dimensions vary between censoring modes and estimation algorithms. Its input may include half ($T=K/2$) or all of the latent features ($T=K$), or it may include one-hot encoded nuisance values ($T=K+|S|$) or task labels ($T=K+|S|+C$). Its output may be a scalar value ($L=1$), or a predicted probability vector over nuisance labels ($L=|S|$).
Sections~\ref{sec:adv}, \ref{sec:dre}, or \ref{sec:wasserstein} explain the structure of this model in more detail as well as how it is applied to the projected features to compute a regularization penalty. The censoring model's parameters are updated in an alternating optimization against the task model.

\subsection{Unregularized and Regularized Training}
In the empirical risk minimization (ERM) framework, the \textit{risk} $R(\theta)$ is defined as the expected loss of a model for a particular set of parameters $\theta$, when applied to samples from a particular dataset and evaluated using a chosen loss function~\cite{vapnik1991principles}. 
For convenience, let $\mathcal{Q}_Y \coloneqq q_{\theta_F, \theta_G}(Y|X)$ represent the label posterior estimated by our task model $G_{\theta_G}(F_{\theta_F}(X))$, 
and let $\mathcal{P}_Y \coloneqq p(Y|X)$ represent the empirical label posterior.
We use the cross-entropy loss $\lossCE$, and thus we can define the empirical risk as:
\begin{align}
    R(\theta_F, \theta_G) = \E_{p(X, Y, S)} \bigg[ \lossCE(\mathcal{P}_Y, \mathcal{Q}_Y) \bigg].
    \label{eq:basic-erm-obj}
\end{align}
To find optimal parameters, we minimize the risk: $\min_{\theta_F, \theta_G,} R(\theta_F, \theta_G)$.

\subsection{Deriving Censoring objectives}
In the censoring framework, we add a regularization term $\lossCensor$
to the optimization problem above. Note this regularization term depends on the parameters $\theta_F$ and $\theta_P$ of the task model's encoder and projector, and the parameters $\theta_J$ of the censoring model. 
To find the optimal regularized parameters, we minimize the regularized risk: 
\begin{align}
 \min_{\theta_F, \theta_G, \theta_P} R(\theta_F, \theta_G) + \lambda \max_{\theta_J} \lossCensor.
    \label{eq:regularized-erm-obj}
\end{align}
The purpose of this regularization term is to help enforce one or more statistical relationships that we expect should hold true, according to the generative model we assume for the task. To obtain a tractable penalty, we must first convert these statistical relationships (such as $Z \perp S$) into concrete quantities that we can estimate or compute analytically, such as a mutual information or a divergence between two distributions. Then, we can create algorithms to estimate these concrete quantities. Finally, we can use these estimates in our regularization objective while training our model.

\paragraph{Using divergences to measure statistical relationships.}
The three censoring modes that we consider each reflect a particular statement about the dependence or independence of variables. 
We consider two concrete quantities that can be used to measure dependence between variables; mutual information (MI) and Wasserstein-1 ($\text{W}_1$) distance.
In general, we can replace a statement about the independence of two variables $A$ and $B$ with a statement about the statistical divergence between the joint distribution $p(A,B)$ and the product of marginal distributions $p(A)p(B)$. 
This comparison is often made using the Kullback-Leibler (KL) divergence, which yields Mutual Information (MI) $I(A;B)$.
We consider several ways to estimate MI in order to enforce independence (and dependence) relationships in Section~\ref{sec:adv} and \ref{sec:dre}.
However, this comparison may also be made using other measures; in Section~\ref{sec:wasserstein}, we replace KL divergence with the Wasserstein-1 ($\text{W}_1$) metric.

\subsection{Adversarial Classifier Baseline}
\label{sec:adv}
As a baseline regularization method, we consider a well-studied approach where the censoring model $J_{\theta_J}(\wc)$ is an adversarial classifier~\cite{wang2018invariant,ozdenizci2019transfer,ozdenizci2019adversarial,ozdenizci2020learning}. 

\paragraph{Marginal Censoring}
Algorithm~\ref{alg:adv_marginal} describes how to compute the regularization penalty in \eqref{eq:regularized-erm-obj} using this adversarial classifier method for the case of marginal censoring.
Recall that, in this case, we seek to enforce $Z \perp S$; to achieve this, we will compute a regularization penalty $\lossCensor$ that approximates $I(Z;S)$.
\begin{algorithm}[tb]
\DontPrintSemicolon
\KwIn{
    Tuples of data, label, nuisance $\{ (x_i, y_i, s_i) \}_{i=1}^{N}$,
    encoder $F_{\theta_F}$, 
    projector $P_{\theta_P}$,
    adversarial classifier $J_{\theta_J}$
}
\KwOutput{
    $\lossCensor$ approximating $I(Z;S)$
}
    \For{$i \in 1\ldots N$}{
        $\tilde{z_i} \gets F_{\theta_F}(x_i)$ \tcp*[r]{Encode}
        $z_i \gets P_{\theta_P}(\tilde{z_i})$ \tcp*[r]{Project}
        $\mathcal{L}_i \gets \lossCE \big( q_{\theta_J}(s_i | z_i), \ p(s_i|z_i) \big)$ \;
    }
    \Return \shortminus $avg( \mathcal{L} )$ \tcp*[r]{Mean CE loss}
\caption{Marginal Censoring using Adversarial Classifier}
\label{alg:adv_marginal}
\end{algorithm}

The adversarial classifier is trained alongside the task model in an alternating optimization scheme; its objective is to use the observed latent features $Z$ to predict the nuisance label $S$. Intuitively, if the MI between these variables is high, the adversary will be able to predict the nuisance label well, and thus the adversary's classification performance can serve as a proxy measure for the mutual information $I(Z;S)$. 

For convenience, here let $\mathcal{Q}_S \coloneqq q_{\theta_J}(S|Z)$ refer to the censoring model's predicted distribution over nuisance labels, and let $\mathcal{P}_S \coloneqq p(S|Z)$ refer to the corresponding ground-truth (one-hot) distribution. We can see that the censoring model's cross-entropy loss $\lossCE\big(\mathcal{P}_S, \ \mathcal{Q}_S \big)$ serves as a lower bound on $I(Z;S)$, as follows.
Note that the MI can be decomposed as $I(Z;S) = H(S) - H(S|Z)$. The marginal entropy $H(S)$ is constant during our optimization process, since it only depends on the data distribution. 
We can obtain a bound on the other term, the conditional entropy $H(S|Z)$, by writing out the definition of cross entropy:
\begin{align}
    \!\!\!\! \lossCE(\mathcal{P}_S, \mathcal{Q}_S)
    & \! = \! H(S|Z) \! + \! \underbrace{\kldiv{\mathcal{P}_S}{\mathcal{Q}_S}}_{\geq 0} \! \geq \! H(S|Z).
\end{align}
Thus we can relate the censoring model's cross-entropy and the MI we seek to minimize:
\begin{align}
    \!\!I(Z;S) & = H(S) \shortminus H(S|Z) \geq H(S) \shortminus \lossCE(\mathcal{P}_S, \mathcal{Q}_S).
\end{align}
In order to enforce $Z \perp S$, we seek to minimize $I(Z;S)$; however if we minimize $\lossCE(\mathcal{P}_S, \mathcal{Q}_S)$ as a proxy, we are actually minimizing a lower bound on the desired quantity. As shown above, this bound will be close when $\kldiv{\mathcal{P}_S}{\mathcal{Q}_S}$ is small, which may occur when the censoring model is sufficiently flexible and trained to convergence.

Training a regularized model using the adversarial classifier involves alternating between updating the parameters of the censoring model using
\begin{align}
    \theta_J^* & = \argmin_{\theta_J} \lossCE(\mathcal{P}_S, \mathcal{Q}_S),
    \label{eq:adv_obj}
\end{align}
and updating the parameters of the task model using \eqref{eq:regularized-erm-obj} where the regularization penalty $\lossCensor$ is obtained using Algorithm~\ref{alg:adv_marginal}.

\paragraph{Conditional and Complementary Censoring}
This method can also be used for conditional censoring. Recall that in conditional censoring we seek to enforce $Z \perp S | Y$.
This corresponds to reducing the conditional MI $I(Z;S|Y) = H(S|Y) - H(S|Z,Y)$.
We can modify the censoring model so that it takes both features and task label as input, and tries to predict the conditional probability over nuisance labels; let $q_{\theta_J}(S|Z,Y)$ represent the output of this modified censoring model.
The first term $H(S|Y)$ is constant with respect to our optimization process; as before, the second term $H(S|Z,Y)$ can be bounded by the cross entropy $\lossCE\big(p(S|Z,Y), \ q_{\theta_J}(S|Z,Y) \big)$ using an analogous derivation. Thus the censoring model's cross-entropy again gives us a bound on the desired MI term.

In the case of complementary censoring, recall that we seek to enforce one independence relationship $Z \perp S$ and one dependence relationship $W \not \perp S$; we achieve this by applying the same censoring model twice. 
For the first set of latent features $Z$, we use the same procedure as in the marginal censoring case; for the second set of latent features $W$, we use the same procedure and invert the sign of the final regularization term. 
This results in an objective of the form 
\begin{align}
    \!\!\!\! \min_{\theta_F, \theta_G, \theta_P} R(\theta_F,\theta_G) + \lambda \max_{\theta_J} (\mathcal{L}_{\textsc{censor,Z}} - \mathcal{L}_{\textsc{censor,W}}), 
\label{eq:adv_comp_obj}
\end{align}
where $\mathcal{L}_{\textsc{censor,Z}}$ regularizes $Z$ and $\mathcal{L}_{\textsc{censor,W}}$ regularizes $W$.

\subsection{Density Ratio Censoring}
\label{sec:dre}
As described above, in the adversarial classifier approach, the adversary's cross-entropy loss provides a lower bound on one or more mutual information terms. 
Here, the censoring model $J_{\theta_J}(\wc)$ is trained to directly estimate the mutual information.

\paragraph{Density Ratio Estimation}
We first briefly introduce a method for density ratio estimation established in the generative modelling literature~\cite{sugiyama2010density}.
Given two distributions over the same space $p(x)$ and $q(x)$, we can estimate the log ratio of their densities $\log \big( p(x) / q(x) \big)$ by training a binary classifier $C$ to distinguish between samples from $p$ versus $q$.
By minimizing the cross-entropy objective,
\begin{align}
\min_{C} \E_{p(x)} \big[ \shortminus \log \sigma (  C(x) ) \big] + \E_{q(x)} \big[ \shortminus \log \sigma ( \shortminus C(x) ) \big],
\label{eq:disc_obj}
\end{align}
where $\sigma(z) = 1 / (1 + e^{-z})$, and $C(x)$ is the logit of the binary classifier, we obtain an optimal classifier $C^*$ whose output is the desired log ratio $C^*(x) = \log \frac{p(x)}{q(x)}$.
In the case of generating synthetic data, the objective in~\eqref{eq:disc_obj} is used to train a discriminator between samples of the true data distribution and the synthetic data distribution~\cite{nguyen2010estimating,nowozin2016f,pu2017adversarial,rhodes2020telescoping}. %

\paragraph{Marginal Censoring}
This density ratio estimation technique can be directly applied for estimating the mutual information between two variables; the censor model $J_{\theta_J}$ plays the role of the binary classifier $C$ above. Algorithm~\ref{alg:wyner_marginal_training} describes how to train this density ratio estimator model.
Recall that mutual information is defined as an expected log-likelihood ratio
\begin{align}
    I(Z;S) := \E_{p(Z, S)} \left[ \log \frac{p(Z,S)}{p(Z)p(S)} \right].
\end{align}
The censoring model's training objective is,
\begin{align}
\min_{\theta_J} 
& \E_{p(Z,S)} [ -\log \sigma (  J_{\theta_J}(Z, S) ) ] \nonumber \\ 
+ & \E_{p(Z) p(S)} [ -\log \sigma ( -J_{\theta_J}(Z, S) ) ],
\label{eq:ratio_obj}
\end{align}
such that $J_{\theta_J}$ learns to approximate $\log \frac{p(Z,S)}{p(Z)p(S)}$.
Note that this training objective requires samples from the empirical joint distribution $p(Z,S)$ as well as from the product of marginal distributions $p(Z) p(S)$. 
Samples from $p(Z) p(S)$ can be approximated by simply permuting one of the variables.
To see that this shuffling gives the desired samples, consider first sampling and encoding a batch of items $\{ Z, Y, S \}$ and discarding $Y, S$. This gives an approximate sample from the marginal distribution $p(Z)$, whose order is unimportant. Likewise sample items from $p(S)$ by discarding $Z,Y$ and optionally shuffling. 
By sampling one batch and only shuffling $S$, we perform these two processes in one step.
\begin{algorithm}[tb]
\DontPrintSemicolon
\KwIn{
    Tuples of data, label, nuisance $\{ (x_i, y_i, s_i) \}_{i=1}^{N}$,
    encoder $F_{\theta_F}$, 
    projector $P_{\theta_P}$,
    density ratio estimator $J_{\theta_J}$
}
\KwOutput{
    Loss for training $\theta_J$
}
    $\tilde{S} \gets permute(S)$ \;
    \For{$i \in 1\ldots N$}{
        $\tilde{z}_i \gets F_{\theta_F}(x_i)$ \tcp*[r]{Encode}
        $z_i \gets P_{\theta_P}( \tilde{z}_i )$ \tcp*[r]{Project}
        $\mathcal{L}_{i}^{\textsc{joint}} \gets -\log \sigma ( J_{\theta_J}(z_i, s_i) )$ \tcp*[r]{$p(Z,S)$}
        $\mathcal{L}_{i}^{\textsc{prod}} \gets -\log \sigma( -J_{\theta_J}(z_i, \tilde{s}_i))$ \!\!\!\! \tcp*[r]{\!$p(Z)p(S)$}
    }
    \Return $avg(\mathcal{L}_{\textsc{joint}}) + avg(\mathcal{L}_{\textsc{prod}})$ \tcp*[r]{Eq~\eqref{eq:ratio_obj}}
\caption{Computing Training Loss for Density Ratio Estimator}
\label{alg:wyner_marginal_training}
\end{algorithm}

The density ratio estimator model can then be used to approximate mutual information as
\begin{align}
I(Z; S) \approx \E_{p(Z,S)} [ J_{\theta_J}(Z, S) ].
\label{eq:wyner_MI_est}
\end{align}
\begin{algorithm}[tb]
\DontPrintSemicolon
\KwIn{
    Tuples of data, label, nuisance $\{ (x_i, y_i, s_i) \}_{i=1}^{N}$,
    encoder $F_{\theta_F}$, 
    projector $P_{\theta_P}$,
    density ratio estimator $J_{\theta_J}$
}
\KwOutput{
    $\lossCensor$ approximating $I(Z;S)$
}
    \For{$i \in 1\ldots N$}{
        $\tilde{z}_i \gets F_{\theta_F} ( x_i )$ \tcp*[r]{Encode}
        $z_i \gets P_{\theta_P}( \tilde{z}_i )$ \tcp*[r]{Project}
        $\mathcal{L}_i \gets J_{\theta_J}(z_i, s_i)$ \tcp*[r]{Eq~\eqref{eq:wyner_MI_est}}
    }
    \Return $avg(\mathcal{L})$
\caption{Marginal Censoring using Density Ratio Estimator}
\label{alg:wyner_marginal_inference}
\end{algorithm}
The overall procedure for training with density ratio censoring involves alternating between updating the parameters $\theta_J$ of the censoring using Algorithm~\ref{alg:wyner_marginal_training}, and updating the parameters of the task model using \eqref{eq:regularized-erm-obj}, where the regularization penalty $\lossCensor$ is given by Algorithm~\ref{alg:wyner_marginal_inference}.

\paragraph{Conditional and Complementary Censoring}
To perform conditional censoring using the density ratio estimation method, we adjust the training objective for $\theta_J$ from \eqref{eq:ratio_obj} as follows.
We seek to enforce the conditional independence $Z \perp S | Y$, which corresponds to minimizing the conditional mutual information $I(Z;S|Y)$.
By chain rule of mutual information, we have $I(Z ; S | Y) = I(Z, Y ; S) - I(Y; S)$.
Since $I(Y; S)$ is fixed with respect to our optimization process, $I(Z, Y ; S)$ is a suitable proxy to minimize.
In order to estimate this quantity, we first adjust the censoring model to accept three inputs instead of two.
The definition of MI states that
\begin{align}
    I(Z,Y;S) := \E_{p(Z,Y,S)} \left[ \log \frac{p(Z,Y,S)}{p(Z,Y)p(S)} \right].
\end{align}
We can estimate the inner log density ratio $\log \frac{p(Z,Y,S)}{p(Z,Y)p(S)}$ by training the censor model with
\begin{align}
    \min_{\theta_J} 
        & \E_{p(Z,Y,S)} [ -\log \sigma (  J_{\theta_J}(Z, Y, S) ) ] \nonumber \\
        + & \E_{p(Z,Y) p(S)} [ -\log \sigma ( -J_{\theta_J}(Z, Y, S) ) ].
\label{eq:ratio_obj_conditional}
\end{align}
This objective requires samples from $p(Z,Y)p(S)$, which we can obtain by shuffling the nuisance labels within a batch (analogous to the shuffling trick for the marginal case).

To perform complementary censoring, we use the marginal censoring approach twice; once to estimate $I(Z;S)$, and a second time to estimate $I(W;S)$. The resulting objective has the same form as the complementary censoring objective in \eqref{eq:adv_comp_obj}.

\subsection{Wasserstein Censoring}
\label{sec:wasserstein}
In the previous two sections, we enforce independence (or dependence) by minimizing (maximizing) an estimate of mutual information.
Here, we replace mutual information with the Wasserstein-1 ($\text{W}_1$) distance between a joint distribution and a product of marginal distributions.

For two variables $A$ and $B$, the chain rule of probability states that the joint distribution can always be expressed as $p(A,B) = p(A) p(B|A)$.
If $A$ and $B$ are independent, then $p(B|A) = p(B)$, and the joint distribution $p(A,B)$ equals the product of marginals $p(A) p(B)$. Whereas mutual information measures the distance between $p(A,B)$ and $p(A)p(B)$ using the KL divergence, any other notion of statistical divergence may be used to similar effect. 
Following previous work in the generative modeling literature, we consider the Wasserstein-1 ($\text{W}_1$) metric; this approach has been previously described as a Wasserstein dependency measure~\cite{ozair2019wasserstein}.

\paragraph{Marginal Censoring.}
To apply this for marginal censoring, we seek to measure the $\text{W}_1$ distance between $p(Z,S)$ and $p(Z)p(S)$.
Under the Kantorovich-Rubinstein duality theorem~\cite{villani2009optimal}, this distance is
\begin{align}
    \! \!\text{W}_1 (r, q ) = 
    \sup_{\| f\|_L \le 1} \E_{r}[f(Z,S)] \shortminus \E_{q}[f(Z, S)], 
    \label{eq:wasserstein_obj}
    \\
    \text{where } r \coloneqq p(Z,S) \nonumber \text{ and } q \coloneqq p(Z)p(S).
\end{align}
Note that the ``critic'' function $f$ has Lipschitz norm bounded by $1$. 
As established in the generative modeling literature, the critic function $f$ can be implemented using be a neural network with an arbitrary Lipschitz constant $K$, giving an estimate of $K \text{W}_1(\cdot, \cdot)$ that suffices in practice for minimizing or maximizing $\text{W}_1(\cdot, \cdot)$\cite{wasserstein-gan}.
We satisfy this requirement in the standard fashion using spectral normalization~\cite{miyato2018spectral} on each layer of the critic network.
Note that~\eqref{eq:wasserstein_obj} requires samples from $p(Z)p(S)$; we use the same trick as in the Section~\ref{sec:dre} of shuffling the nuisance variable within a batch to obtain such samples.
\begin{algorithm}[tb]
\DontPrintSemicolon
\KwIn{
    Tuples of data, label, nuisance $\{ (x_i, y_i, s_i) \}_{i=1}^{N}$,
    encoder $F_{\theta_F}$, 
    projector $P_{\theta_P}$,
    Wasserstein critic $J_{\theta_J}$
}
\KwOutput{
    $\lossCensor$ approximating $\text{W}_1\big(p(Z,S), \ p(Z)p(S) \big)$
}
    $\tilde{S} \gets permute(S)$ \;
    \For{$i \in 1\ldots N$}{
        $\tilde{z}_i \gets F_{\theta_F} ( x_i )$ \tcp*[r]{Encode}
        $z_i \gets P_{\theta_P}( \tilde{z}_i )$ \tcp*[r]{Project}
        $\mathcal{L}_i^{\textsc{joint}} \gets J_{\theta_J}(z_i, s_i)$ \tcp*[r]{$p(Z,S)$}
        $\mathcal{L}_i^{\textsc{prod}} \gets J_{\theta_J}(z_i, \tilde{s}_i)$ \tcp*[r]{$p(Z)p(S)$}
    }
    \Return $avg(\mathcal{L}^{\textsc{joint}}) - avg(\mathcal{L}^{\textsc{prod}})$
\caption{Marginal Censoring using Wasserstein Critic}
\label{alg:wasserstein_marginal}
\end{algorithm}
Algorithm~\ref{alg:wasserstein_marginal} describes how we can use a critic neural network to estimate the Wasserstein distance in~\eqref{eq:wasserstein_obj} in order to perform marginal censoring.
Note that the critic model receives two inputs.
Training a model using Wasserstein censoring involves alternating between updates to the parameters of the critic model $\theta_J$ and the parameters of the task model; when updating the task model, the output from Algorithm~\ref{alg:wasserstein_marginal} is used directly; when updating the critic model, the same loss is used with the sign flipped.

\paragraph{Conditional and Complementary Censoring}
To perform conditional censoring using the Wasserstein method, we adjust the training scheme described above as follows.
First, the critic model is adjusted to accept three inputs (observed features $Z$, task labels $Y$, and nuisance labels $S$).
Next, we begin with the same logic as in the case of conditional censoring using the density ratio estimator method (see Section~\ref{sec:dre}).
For that method, we showed that the conditional independence $Z \perp S | Y$ can be enforced by minimizing $I(Z;S|Y)$, and in turn this can be replaced by minimizing $I(Z,Y;S)$. We used this final quantity because we can easily obtain samples from the relevant distributions ($p(Z,Y,S)$ and $p(Z,Y)p(S)$).
Here, we replace the use of KL divergence in $I(Z,Y;S)$ with $\text{W}_1\big( p(Z,Y,S), \ p(Z,Y) p(S) \big)$, which we estimate using a critic neural network as in Algorithm~\ref{alg:wasserstein_marginal} and Equation~\eqref{alg:wasserstein_marginal}.

In complementary censoring, the Wasserstein critic is used twice, once to minimize $\text{W}_1(p(Z,S), p(Z)p(S)$, and once to maximize $\text{W}_1(p(W,S), p(W)p(S))$.

\subsection{Computational Experiments}
\paragraph{Dataset}
We use a large publicly-available EEG dataset for all experiments~\cite{zhang2020benchmark}. 
This dataset contains EEG recordings during a rapid serial visual presentation (RSVP) task with binary trials.
Subjects were presented with a sequence of quickly flashed images and asked to watch for target images, while their EEG responses were recorded. 
Each stimulus presentation is associated with a binary label. 
Data were recorded at $1000$Hz and made available at a down-sampled rate of $250$Hz. 
The dataset includes just over 1 million binary trials, collected from $64$ subjects, each of whom participated in $2$ recording sessions.

\paragraph{Experimental Setup}
In each experiment, we evaluated the performance of a single proposed regularized training method, defined by the parameters listed in Table~\ref{tab:hyperparams}.
The test performance of the regularized model was compared to the test performance of the same model without regularization.
In one half of experiments, models were trained for a fixed number of epochs using all sessions of data from $28$ subjects for training, and using all sessions of data from $4$ subjects for testing.
In the other half of experiments, models were trained with all sessions of data from $24$ subjects for training, $4$ subjects for validation, and $4$ subjects for testing; the model checkpoint from the epoch of best validation performance was used for testing.

We used cross-validation to obtain reliable estimates of model performance. 
Each experiment was repeated $100$ times using $10$ different initial random seeds and $10$ different choices of train/val/test subject assignment.
Note that the dataset contains $64$ total subjects, while each experiment used $32$ subjects; thus the $10$ subject splits are partially overlapping.
Model performance was quantified using balanced accuracy, which is the average of accuracy on each class.

In addition to the data $X$ and binary task labels $Y$, experiments require a nuisance label $S$, computed as an integer that uniquely identifies a particular subject and session. 
Non-target trials were subsampled to achieve a proportion of $10$ non-target trials per $1$ target trial to be similar to real-world RSVP applications such as assistive typing.

\paragraph{Hyperparameters Explored}
Table~\ref{tab:hyperparams} summarizes the hyperparameters varied across experiments.
\begin{table}[htb]
\centering
\begin{tabular}{cc}
\hline
\textbf{Hyperparameter} & \textbf{Range Explored} \\ \hline \hline
\multirow{3}{*}{Censor Mode} & Marginal, \\ 
    & Conditional, \\
    & Complementary \\ \hline
\multirow{3}{*}{Censor Method} & Adversarial Classifier, \\
    & Density Ratio Estimator, \\
    & Wasserstein Critic \\ \hline
\multirow{3}{*}{Censor Strength ($\lambda$)} & 
    0.01, 0.02, 0.03, 0.05, 0.1, \\ %
    & 0.2, 0.3, 0.5, 1, 2, 3, 5, \\ %
    & 10, 20, 30, 50, 100.0 \\ \hline  %
\multirow{2}{*}{Projection Type} & Trivial ($P_{\theta_P} = I$), \\
    & Non-trivial \\ \hline
\multirow{2}{*}{Evaluation Point} & Final Checkpoint, \\
    & Best Val Checkpoint \\ \hline
\end{tabular}
\caption{Hyperparameters varied across experiments. Each experiment was repeated $100$ times, using $10$ random seeds and $10$ splits of train, validation, and test subjects.
\textit{Censor Mode}: choice of graphical model and statistical relationship to enforce (see Section~\ref{sec:graphical-models-censor-modes}).
\textit{Censor Method}: technique used to compute regularization penalty (see Sections~\ref{sec:adv}, \ref{sec:dre}, and \ref{sec:wasserstein}).
\textit{Projection Type}: whether projection network $P_{\theta_P}$ is the identity function (see Figure~\ref{fig:model-schematic}.
\textit{Censor Strength}: value of coefficient $\lambda$ in \eqref{eq:regularized-erm-obj}.
\textit{Evaluation Point}: whether model is evaluated at epoch of best validation accuracy, or final ($100$th) epoch. 
}
\label{tab:hyperparams}
\end{table}

For marginal and conditional censoring, the dimension of $\tilde{Z}$ and $Z$ was $128$.
For complementary censoring, $64$ dimension were used for $Z$ and $64$ for $W$.
Models were implemented and trained using PyTorch~\cite{paszke2019pytorch} and Pytorch Lightning~\cite{pytorch_lightning}, using the AdamW optimizer~\cite{loshchilov2017decoupled} with constant learning rate $10^{-4}$, default values of $\beta_1=0.9, \beta_2=0.999$, and batch size $1024$. 
For experiments that evaluate the final model checkpoint, training lasted $100$ epochs. 
For experiments that evaluate the best validation checkpoint, training lasted up to $30$ epochs (since the point of optimal early stopping almost always occurs before this).

The encoder was a 1D convolutional network with $248$K parameters $\theta_F$.
The classifier was a multi-layer perceptron (MLP) with $50$K parameters $\theta_G$.
When present, the projection network was an MLP with $66$K parameters $\theta_P$.
The censoring model was an MLP, with between $48$K and $56$K parameters $\theta_J$, depending on the number of input vectors ($Z,S$ for marginal censoring, $Z,Y,S$ for conditional, $Z,Y$ and $W,Y$ with $dim(Z)=dim(W)=64$ for complementary) and the dimension of the output ($dim(S)$ for the adversarial classifier method; $1$D for the density ratio estimation and Wasserstein censoring methods). 

\paragraph{Cross-Validation}
To obtain a stable estimate of the effect of our proposed methods, each experiment was run $100$ times, using $10$ cross-validation folds for each of $10$ random seeds.
This helped control for variation due to the particular assignment of subjects into train, validation, and test sets, as well as variation due to weight initialization and batch selection during training.
Note that in a single cross-validation fold, the subjects used for train, validation, and test are all disjoint.

\section{Results}
\begin{figure*}[htb]
    \centering
    \begin{subfigure}[b]{0.8\linewidth}
        \includegraphics[width=\linewidth]{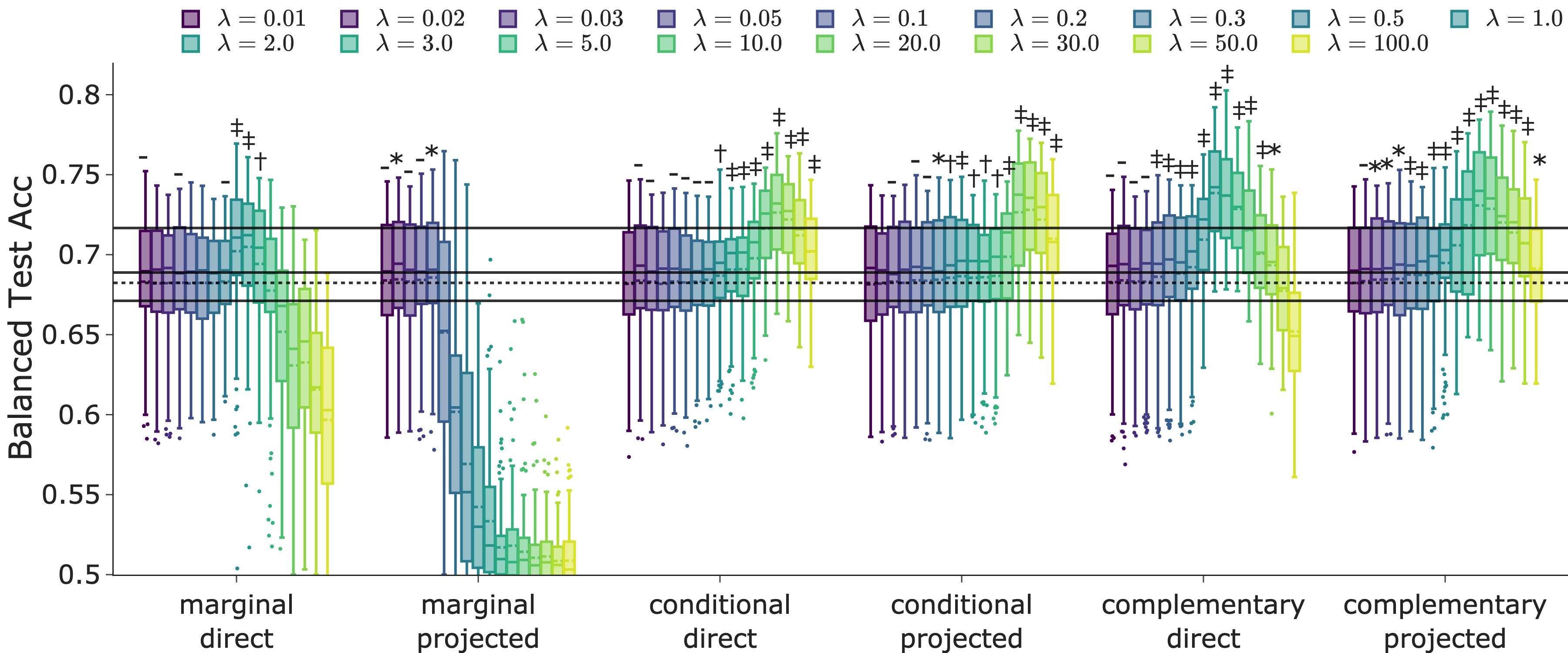}
        \vspace{-\baselineskip}
        \caption{Adversarial Censoring (Baseline)}
        \label{fig:boxplots-final-ckpt-adv}
        \vspace{\baselineskip}
    \end{subfigure}
    \begin{subfigure}[b]{0.8\linewidth}
        \includegraphics[width=\linewidth]{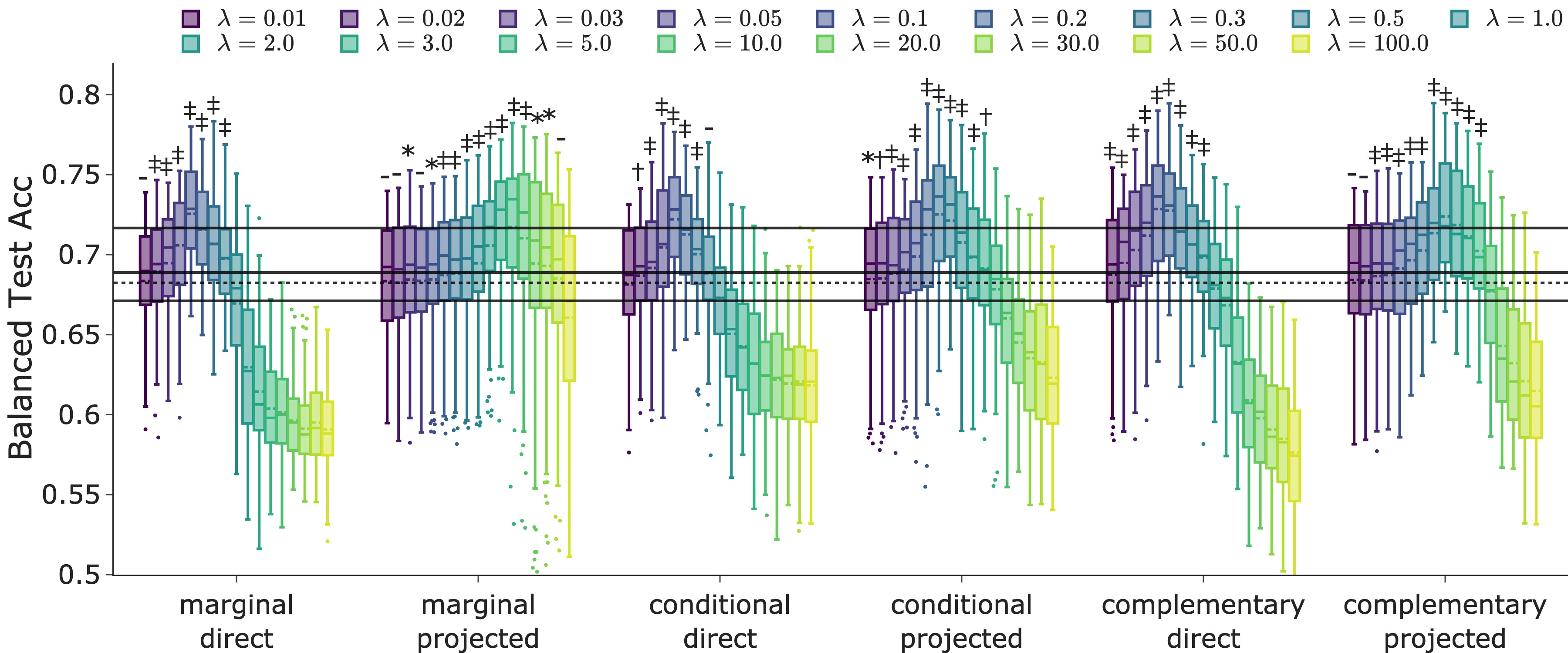}
        \vspace{-\baselineskip}
        \caption{Density Ratio Censoring}
        \label{fig:boxplots-final-ckpt-dre}
        \vspace{\baselineskip}
    \end{subfigure}
    \begin{subfigure}[b]{0.8\linewidth}
        \includegraphics[width=\linewidth]{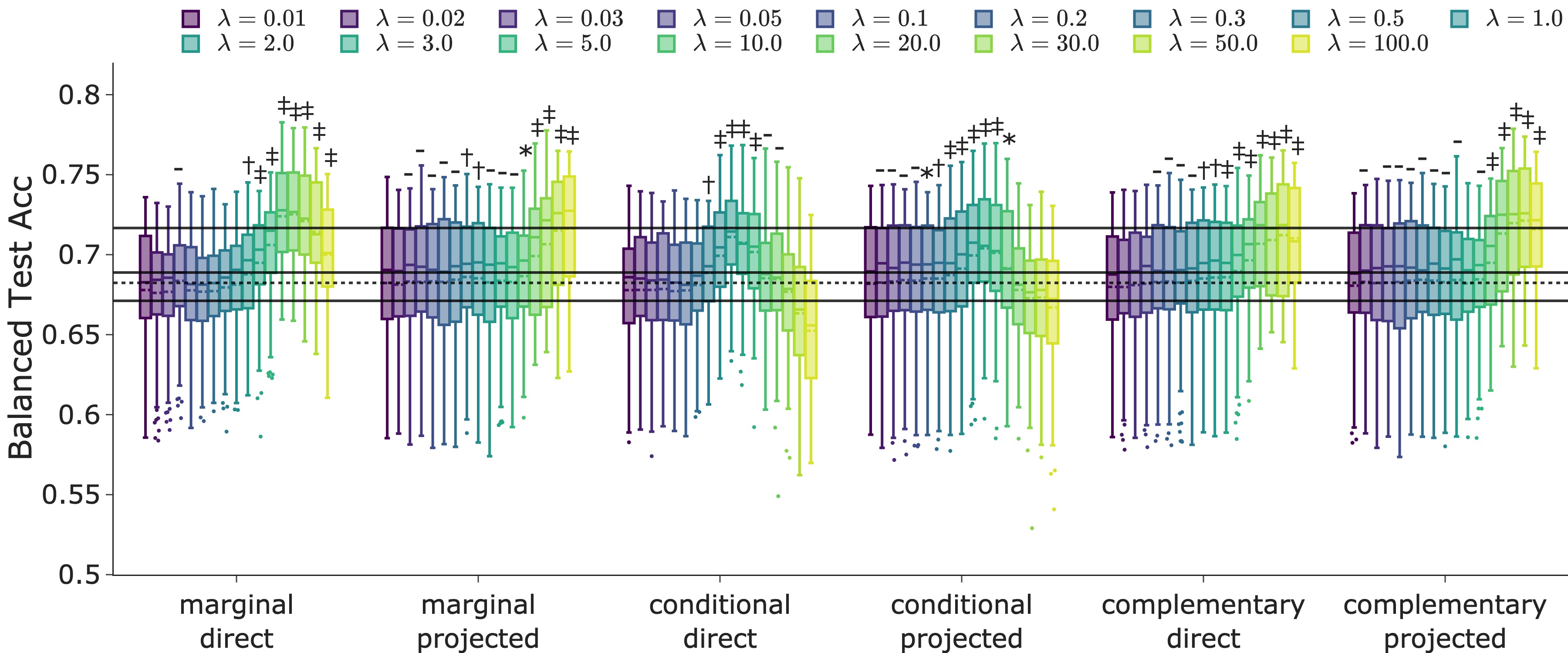}
        \vspace{-\baselineskip}
        \caption{Wasserstein Censoring}
        \label{fig:boxplots-final-ckpt-was}
    \end{subfigure}
    \caption{
    Balanced test accuracy of different regularization strategies when models are trained for a fixed number of epochs.
    \subref{fig:boxplots-final-ckpt-adv}: adversarial classifier baseline (Sec~\ref{sec:adv}). 
    \subref{fig:boxplots-final-ckpt-dre}: density ratio censoring (Sec~\ref{sec:dre}). 
    \subref{fig:boxplots-final-ckpt-was}: Wasserstein censoring (Sec~\ref{sec:wasserstein}). 
    Boxplots show $100$ trials, varying random seed and data split.
    Horizontal black lines show unregularized model performance.
    \textit{Marginal}, \textit{conditional}, \textit{complementary}: censoring modes (Sec~\ref{sec:graphical-models-censor-modes}).
    \textit{Projected}: projection model $P_{\theta_P}$ is non-trivial; \textit{direct}: $P_{\theta_P}$ is omitted.
    $\lambda$: strength of regularization in \eqref{eq:regularized-erm-obj}.
    }
    \label{fig:boxplots-final-ckpt}
\end{figure*}

Figure~\ref{fig:boxplots-final-ckpt} shows the distribution of balanced accuracy on the test set when models were trained for a fixed number of epochs.
The top panel shows the baseline adversarial classifier method, the middle panel shows the proposed density ratio censoring method, and the bottom panel shows the proposed Wasserstein censoring method.
In each panel, a group of boxplots on the X-axis represents a single choice of censoring mode and projection type (e.g. marginal censoring with a trivial projection).
Each single boxplot represents a single value of $\lambda$ in \eqref{eq:regularized-erm-obj}, and shows $100$ repetitions of the experiment across different data folds and random seeds.
The unregularized model's performance is shown by horizontal black lines; solid lines show lower quartile, median, and upper quartile, while the dashed line shows the mean. 
A paired t-test was performed between the $100$ balanced accuracy scores of each censored model and the $100$ scores of the unregularized model; models with a t-statistic greater than zero are annotated (-, $p > 0.05$; *, $0.01 < p \leq 0.05$; $\dagger$, $0.001 < p \leq 0.01$; $\ddagger$, $p \leq 0.001$).

The proposed methods (middle and bottom panel) show a strong benefit over the unregularized model across all censoring modes.
Within each censoring mode, the proposed methods show a benefit at the highest significance level ($\ddagger$) across a wide range of $\lambda$ values.
By contrast, the baseline adversarial classifier method (top panel) shows a reduced benefit, and shows benefit for fewer experimental settings.
Note that we do not perform any correction for multiple hypothesis testing; since tests we perform are not independent, optimal correction is non-trivial. 
Nonetheless, the highest significance level considered ($\ddagger, p < 0.001$) still corresponds to a significant improvement with $p < 0.05$ using a Bonferroni correction across the $17$ values of $\lambda$ in each experimental group.

\begin{figure*}[htb]
    \centering
    \begin{subfigure}[b]{0.8\linewidth}
        \includegraphics[width=\linewidth]{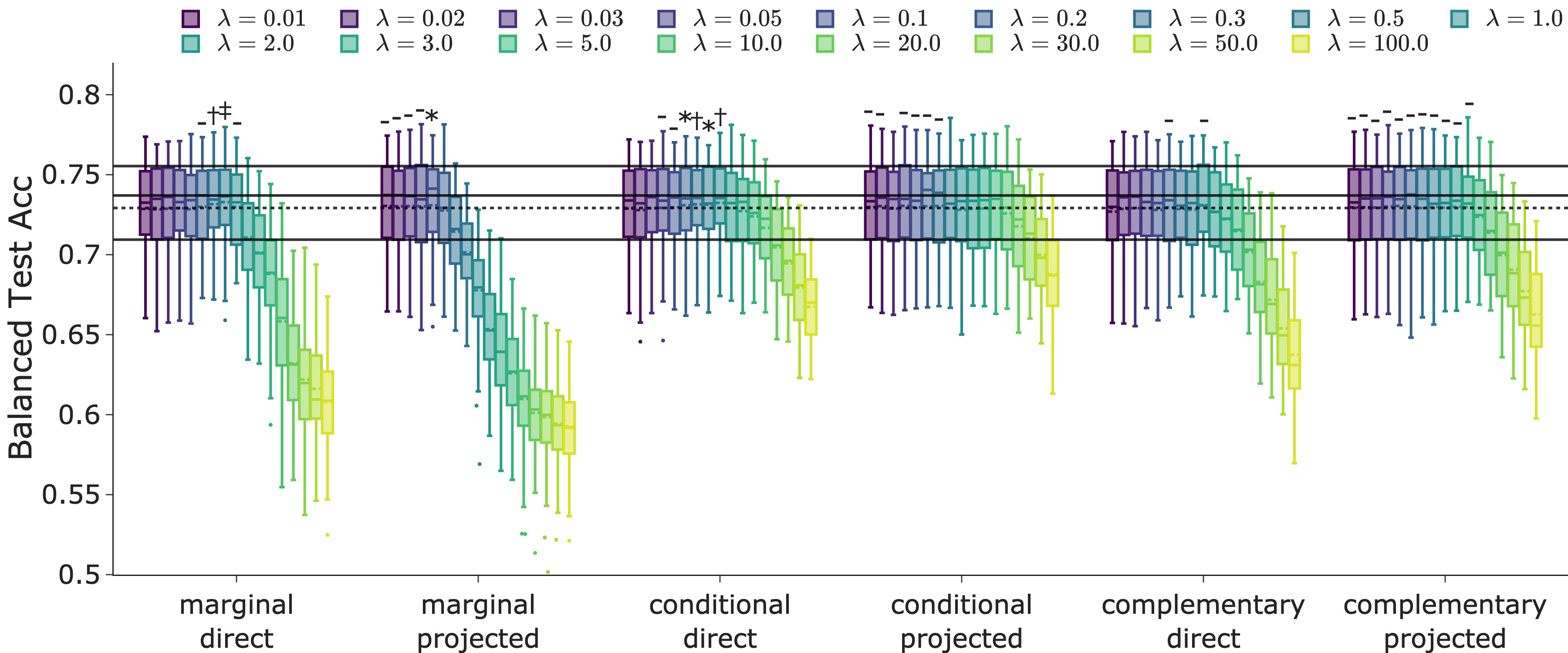}
        \vspace{-\baselineskip}
        \caption{Adversarial Censoring (Baseline)}
        \label{fig:boxplots-best-val-ckpt-adv}
        \vspace{\baselineskip}
    \end{subfigure}
    \begin{subfigure}[b]{0.8\linewidth}
        \includegraphics[width=\textwidth]{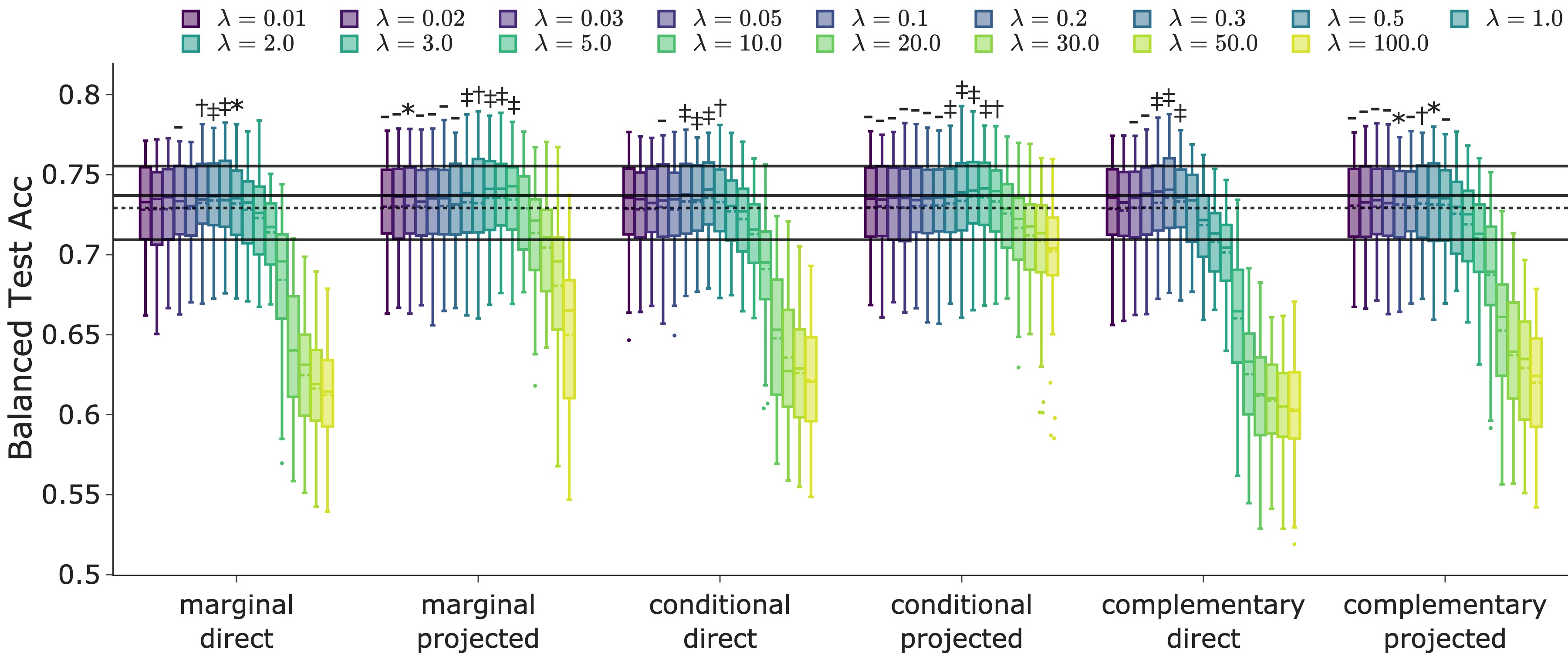}
        \vspace{-\baselineskip}
        \caption{Density Ratio Censoring}
        \label{fig:boxplots-best-val-ckpt-dre}
        \vspace{\baselineskip}
    \end{subfigure}
    \begin{subfigure}[b]{0.8\linewidth}
        \includegraphics[width=\textwidth]{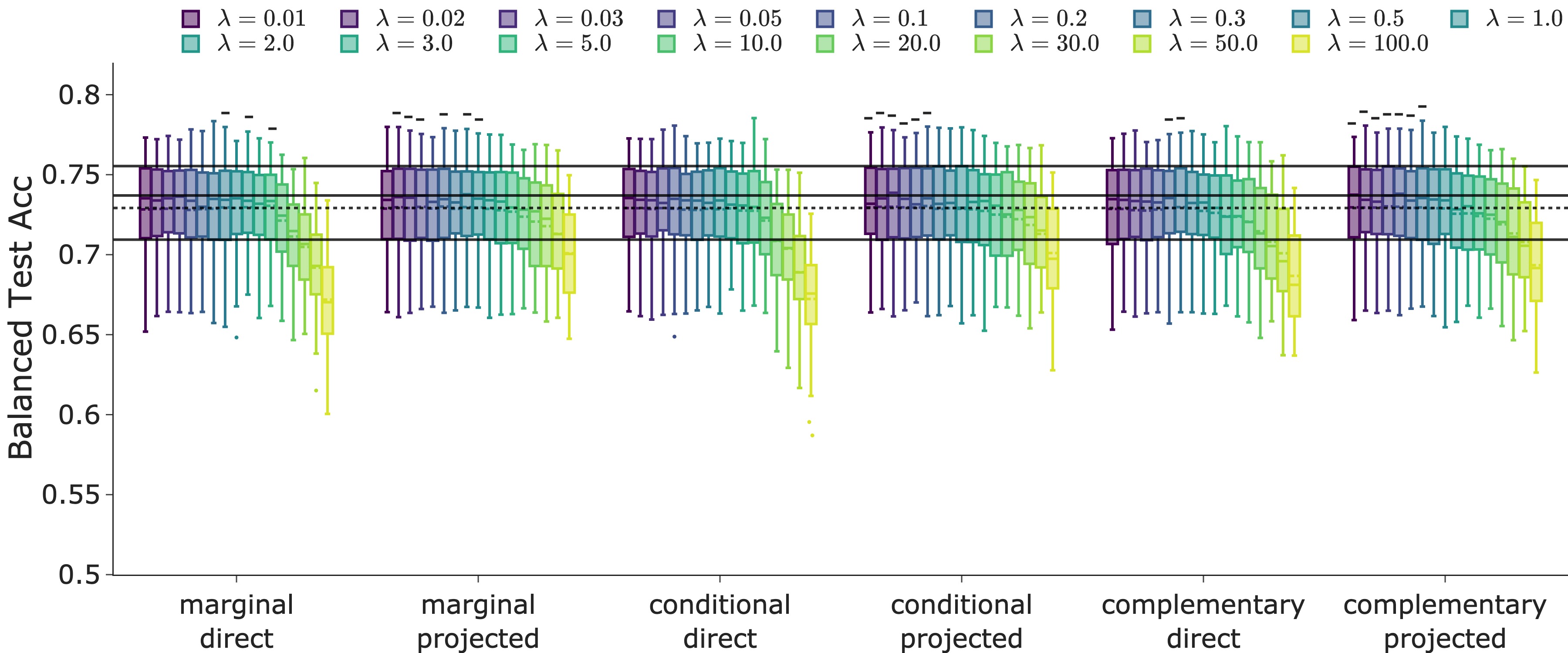}
        \vspace{-\baselineskip}
        \caption{Wasserstein Censoring}
        \label{fig:boxplots-best-val-ckpt-wasserstein}
    \end{subfigure}
    \caption{
    Balanced test accuracy of different regularization strategies when optimal early stopping is also performed.
    \subref{fig:boxplots-best-val-ckpt-adv}: adversarial classifier baseline (Sec~\ref{sec:adv}). 
    \subref{fig:boxplots-best-val-ckpt-dre}: density ratio censoring (Sec~\ref{sec:dre}). 
    \subref{fig:boxplots-best-val-ckpt-wasserstein}: Wasserstein censoring (Sec~\ref{sec:wasserstein}). 
    Boxplots show $100$ trials, varying random seed and data split.
    Horizontal black lines show model performance with early-stopping and no censoring.
    \textit{Marginal}, \textit{conditional}, \textit{complementary}: censoring modes (Sec~\ref{sec:graphical-models-censor-modes}).
    \textit{Projected}: projection model $P_{\theta_P}$ is used; \textit{direct}: $P_{\theta_P}$ is omitted.
    $\lambda$: strength of regularization in \eqref{eq:regularized-erm-obj}.
    }
    \label{fig:boxplots-best-val-ckpt}
\end{figure*}

Figure~\ref{fig:boxplots-best-val-ckpt} shows results analogous to Figure~\ref{fig:boxplots-final-ckpt}, but models were tested using the checkpoint of best validation performance. 
This optimal early stopping already provides a strong regularization to both unregularized and censored models, reducing the potential incremental benefit of censoring.
Note that early stopping requires allocating a subset of training data as a held-out validation set; for some applications, this may not be feasible.

While the benefit of censoring regularization was reduced in these experiments, we observed two important benefits of our proposed methods over the adversarial classifier baseline method.
First, the adversarial classifier only improves upon the uncensored early-stopped model in a narrow subset of settings, while the proposed density ratio method provides a stronger benefit over the uncensored early-stopped model for a larger range of settings.
Second, while the proposed Wasserstein method does not achieve statistically significant benefits over the uncensored early-stopped model, the performance is more consistent across the range of settings explored, indicating that this method is relatively safe to apply even when the optimal hyperparameters are not known.

\begin{figure}
    \centering
    \begin{subfigure}[b]{0.7\linewidth}
        \includegraphics[width=\linewidth]{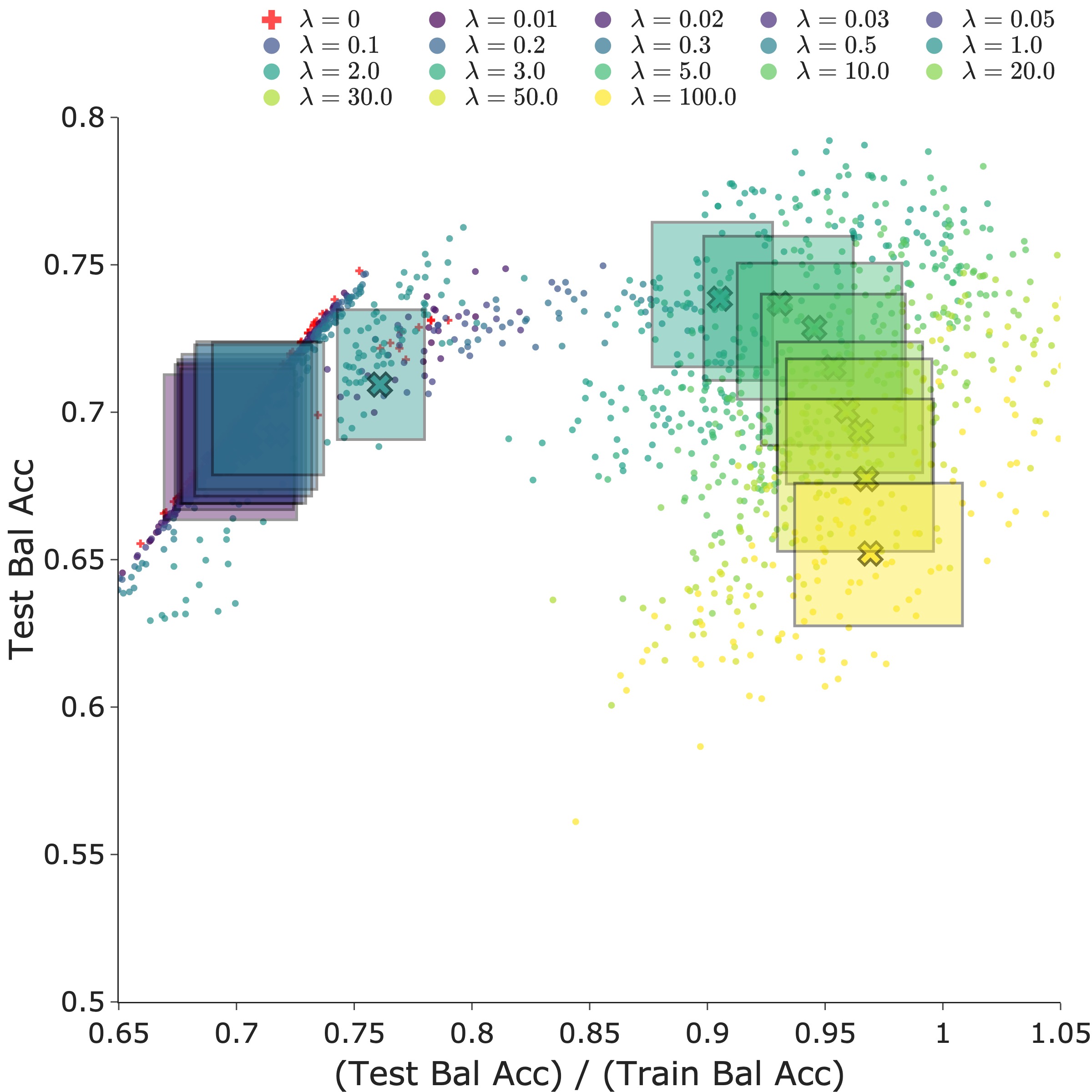}
        \vspace{-\baselineskip}
        \caption{Adversarial Censoring (Baseline)} %
        \label{fig:test-vs-gen-final-ckpt-adv-comp-direct}
        \vspace{\baselineskip}
    \end{subfigure}
    \begin{subfigure}[b]{0.7\linewidth}
        \includegraphics[width=\linewidth]{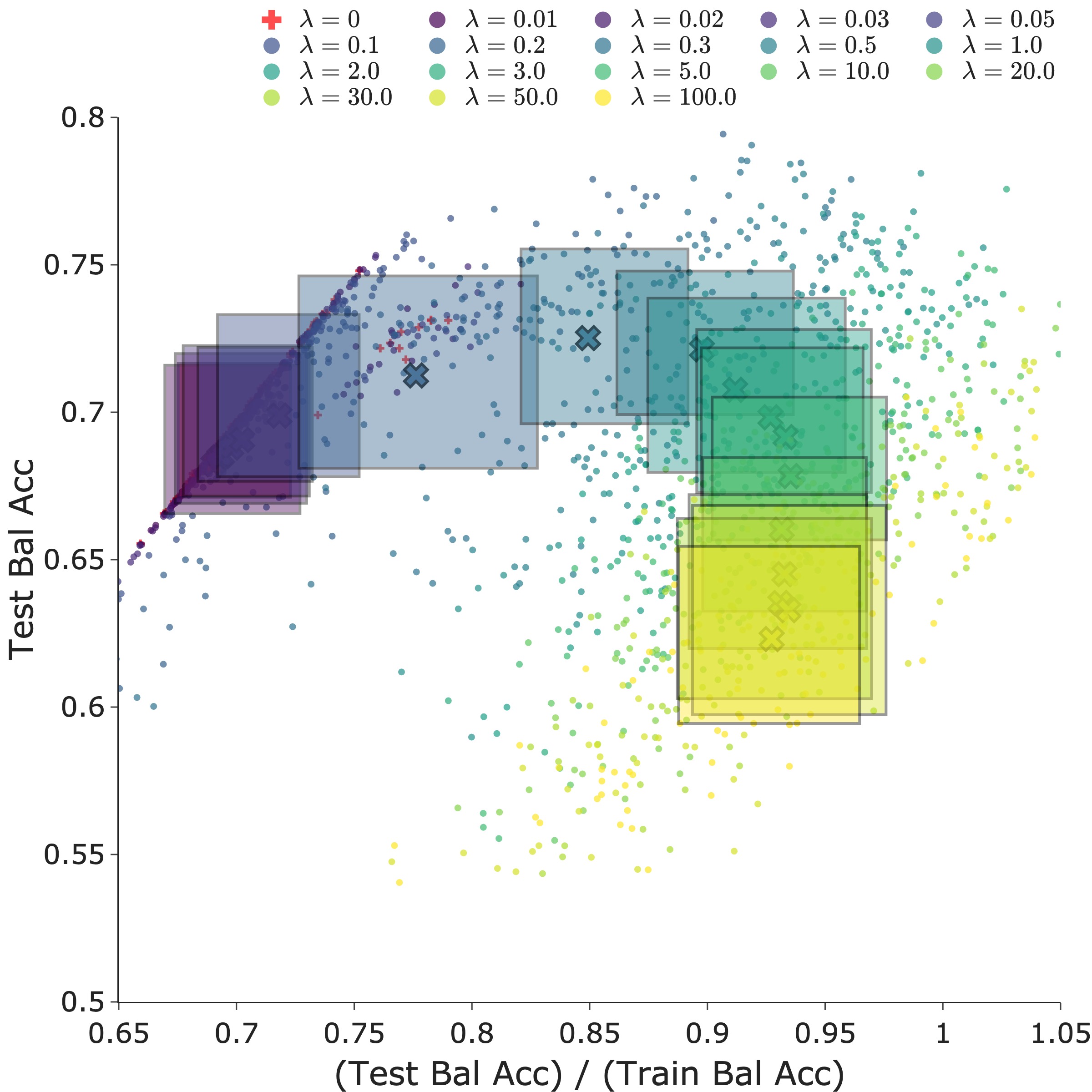}
        \vspace{-\baselineskip}
        \caption{Density Ratio Censoring} %
        \label{fig:test-vs-gen-final-ckpt-dre-cond-proj}
        \vspace{\baselineskip}
    \end{subfigure}
    \begin{subfigure}[b]{0.7\linewidth}
        \includegraphics[width=\linewidth]{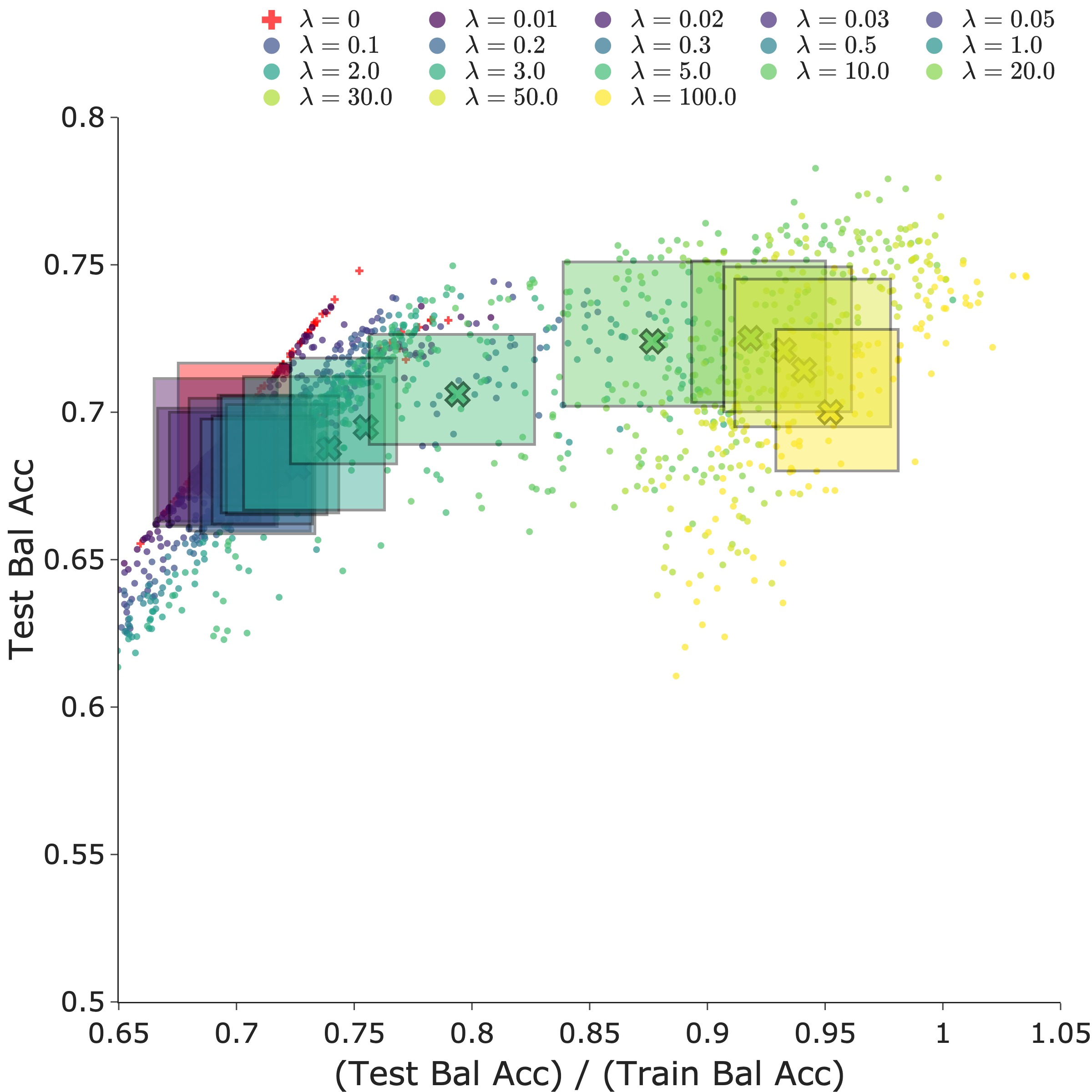}
        \vspace{-\baselineskip}
        \caption{Wasserstein Censoring} %
        \label{fig:test-vs-gen-final-ckpt-was-marg-direct}
    \end{subfigure}
    \caption{Test Balanced Accuracy vs Overfitting ratio; models trained for fixed number of epochs (ideal models have large x and y coordinate).
    \subref{fig:test-vs-gen-final-ckpt-adv-comp-direct}: adversarial censoring, complementary mode, without $P_{\theta_P}$.
    \subref{fig:test-vs-gen-final-ckpt-dre-cond-proj}: density ratio censoring, conditional mode, using $P_{\theta_P}$.
    \subref{fig:test-vs-gen-final-ckpt-was-marg-direct}: Wasserstein censoring, marginal mode, without $P_{\theta_P}$.
    Points show $100$ trials, varying random seed and data split.
    Boxes show interquartile range on each axis.
    }
    \label{fig:test-vs-gen-final-ckpt}
\end{figure}

In addition to increasing test accuracy, the proposed censoring regularization methods also reduced the amount of overfitting. Overfitting was quantified by computing the ratio of balanced accuracy on test data to balanced accuracy on train data.
Figure~\ref{fig:test-vs-gen-final-ckpt} shows plots of balanced test accuracy on the vertical axis, and overfitting ratio on the horizontal axis. Due to space constraints, results of only a few selected model hyperparameters are shown.
An ideal model would have a large y-axis value, indicating strong test performance, and a large x-axis value, indicating that it retains its training performance when transferring to unseen test subjects. Each plot shows the performance of a single censoring method, mode, and choice of projection type, for various values of $\lambda$. 
For each value of $\lambda$, colored points show the $100$ independent reruns across data folds and random seeds; the colored box represents the interquartile range (IQR) of the points along each axis.
The unregularized model ($\lambda = 0$) is also shown in each plot.

\section{Discussion}
We study the problem of regularized model training to perform zero-shot subject transfer learning for EEG classification tasks. Models are trained using a standard cross-entropy loss and a censoring regularization term, with the aim of improving performance on unseen subjects and reducing the gap between train and test performance.

We provide a novel motivation for the censoring regularization strategy. Two assumptions must be met for classifier models to achieve high performance: the dataset being used for training must match the assumed generative model for the task, and the classifier model must learn the dependency structure implied by this generative model. When we observe low model performance, this might occur because one or both of these assumptions is violated. We provide regularization penalties to address the second source of error. Specifically, for any particular generative model, we select a statistical relationship that should hold, convert this to a divergence that should be minimized (here, a mutual information term or a Wasserstein distance), and then add this as a regularization term in the training objective.

By considering several graphical models and their conditional independence structure, we identify three different statistical relationships that can be enforced to regularize the model.
We refer to the choice of a statistical relationship to enforce as a ``censoring mode.'' 
In \textit{marginal} censoring, we enforce marginal independence between the latent features $Z$ and the nuisance labels $S$.
In \textit{conditional} censoring, we enforce conditional independence between the latent features $Z$ and the nuisance labels $S$ given the task labels $Y$.
In \textit{complementary} censoring, we enforce independence between the nuisance labels $S$ and one set of latent features $Z$, while enforcing dependence between $S$ and another set of latent features $W$.

In order to construct a regularization penalty to enforce one these statistical relationships, we must select a quantitative measure of dependency that can be directly optimized, as well as a method for estimating this quantity.
We propose two new quantities and provide simple techniques for estimating them.
In one technique, we use density ratio estimation to compute an approximation of mutual information; 
in the other, we use Wasserstein distance to compute a surrogate for mutual information.
We compare these proposed techniques to a baseline method that uses an adversarial classifier to compute a proxy for mutual information.
These estimation techniques are generic; they can be used for any of the censoring modes discussed above, as well as future censoring modes not considered here.

We evaluated the performance of the proposed estimation methods using extensive computational experiments on a large benchmark EEG dataset. In each experiment, we selected a single censoring mode and estimation technique, and set the strength of the regularization via the coefficient $\lambda$. 
Models were trained with or without censoring regularization.
In some experiments a set of unseen validation subjects was used to perform early stopping, while in other experiments models were trained for a fixed duration. 
We then evaluated the balanced accuracy of the model on unseen test subjects.
By varying random seed and the split of data into train, validation, and test sets, we characterized the distribution of model performance.

We found that, when evaluated at the final epoch of training, our techniques provide a significant increase in test accuracy for all censoring modes across a wide range of hyperparameter values, as well as greatly decreasing the gap between test and train performance. Compared to the baseline adversarial classifier technique, the proposed methods were stronger and more stable, giving a greater benefit in test performance across a wider range of hyperparameter values.

We also evaluated performance at the epoch of best validation performance, in order to understand whether the benefit of censoring regularization is redundant with the benefits of early stopping. Early stopping is a well-studied and widely applicable technique for regularization, but requires allocating a portion of training data for validation, which may undesirable or infeasible in applications with limited data. 
We found that the benefit of our density ratio estimation technique was reduced when used alongside early stopping, but still statistically significant in some cases. This indicates that our method provides regularization that is complementary to the benefits of early stopping. 
By comparison, the adversarial classifier baseline's benefit was further reduced. The Wasserstein critic method's benefit was also greatly reduced, but it had the advantage of giving consistent performance across censoring modes and $\lambda$ values, whereas the adversarial baseline became highly sensitive to these hyperparameters.

Our may be extended by considering other possible generative models, and selecting one or multiple other statistical relationships to enforce during training. Our techniques could also be adapted to use other quantitative measures of statistical dependence. When comparing the joint distribution and product of marginals as a means of estimating dependence between two variables, we considered KL divergence (leading to the mutual information measure) and Wasserstein-1 distance (leading to the Wasserstein critic technique); any other measure of statistical distance or divergence would also be suitable, such as other $f$-divergences~\cite{renyi1961measures,rubenstein2019practical,sreekumar2022neural}, Maximum Mean Discrepancy~\cite{gretton2012kernel}, or other methods for estimating Wasserstein distance and related measures such as Sinkhorn divergences~\cite{genevay2018learning}.

\printbibliography

\end{document}